\documentclass[runningheads]{llncs}

\usepackage{eccv}
\usepackage{eccvabbrv}
\usepackage{float}
\usepackage{graphicx}
\usepackage{booktabs}
\usepackage{bm}
\usepackage[accsupp]{axessibility}  
\usepackage{hyperref}

\usepackage{orcidlink}
\usepackage{algorithm,algpseudocode}
\usepackage{lipsum}

\makeatletter
\newenvironment{breakablealgorithm}
  {
   \begin{center}
     \refstepcounter{algorithm}
     \hrule height.8pt depth0pt \kern2pt
     \renewcommand{\caption}[2][\relax]{
       {\raggedright\textbf{\ALG@name~\thealgorithm} ##2\par}%
       \ifx\relax##1\relax 
         \addcontentsline{loa}{algorithm}{\protect\numberline{\thealgorithm}##2}%
       \else 
         \addcontentsline{loa}{algorithm}{\protect\numberline{\thealgorithm}##1}%
       \fi
       \kern2pt\hrule\kern2pt
     }
  }{
     \kern2pt\hrule\relax
   \end{center}
  }
\makeatother

\begin{document}

\title{On Learning Discriminative Features from Synthesized Data for Self-Supervised Fine-Grained Visual Recognition} 

\titlerunning{Self-Supervised Fine-Grained Visual Recognition}

\author{Zihu Wang\inst{1}\orcidlink{0009-0004-5132-2613} \and
Lingqiao Liu\inst{2}\orcidlink{0000-0003-3584-795X} \and
Scott Ricardo Figueroa Weston\inst{1} \and
Samuel Tian\inst{3} \and
Peng Li\inst{1}\orcidlink{0000-0003-3548-4589}
}

\authorrunning{Z.~Wang et al.}

\institute{University of California, Santa Barbara CA 93106, USA \and
The University of Adelaide, Adelaide South Australia 5001, Australia \and Carnegie Mellon University, Pittsburgh PA 15213, USA 
\email{\{zihu\_wang,scottricardo,lip\}@ucsb.edu}
\email{\{lingqiao.liu\}@adelaide.edu.au}
\email{\{samuel.tian7\}@gmail.com}
}

\maketitle

\begin{abstract}

Self-Supervised Learning (SSL) has become a prominent approach for acquiring visual representations across various tasks, yet its application in fine-grained visual recognition (FGVR) is challenged by the intricate task of distinguishing subtle differences between categories. To overcome this, we introduce an novel strategy that boosts SSL's ability to extract critical discriminative features vital for FGVR. This approach creates synthesized data pairs to guide the model to focus on discriminative features critical for FGVR during SSL. We start by identifying non-discriminative features using two main criteria: features with low variance that fail to effectively separate data and those deemed less important by Grad-CAM induced from the SSL loss. We then introduce perturbations to these non-discriminative features while preserving discriminative ones. A decoder is employed to reconstruct images from both perturbed and original feature vectors to create data pairs. An encoder is trained on such generated data pairs to become invariant to variations in non-discriminative dimensions while focusing on discriminative features, thereby improving the model's performance in FGVR tasks. We demonstrate the promising FGVR performance of the proposed approach through extensive evaluation on a wide variety of datasets.

\end{abstract}

\section{Introduction}
\label{sec:intro}
In computer vision, Fine-grained Visual Recognition (FGVR) focuses on identifying subcategories of visual data, such as bird species \cite{berg2014birdsnap,wah2011caltech}, aircraft variants \cite{maji2013fine}, and vehicle models\cite{krause20133d}. Different from studies on large-scaled general image datasets \cite{ILSVRC15,thomee2016yfcc100m,krizhevsky2009learning}, FGVR tasks highlight the challenge of distinguishing subtle visual patterns.

Self-Supervised Learning (SSL) methods have recently largely advanced the domain of visual representation learning, circumventing the necessity for human-provided annotations. In SSL, many contrastive learning approaches achieve state-of-the-art performance by learning the similarities between data pairs derived from augmentations of identical source images \cite{chen2020simple,he2020momentum,chen2021exploring,caron2021emerging,wang2023contrastive,grill2020bootstrap,lee2021improving}. These methods facilitate the transferability of the learned representations across a wide range of visual recognition problems \cite{chen2020simple,he2020momentum,gao2022disco,ericsson2021well}. Despite these advancements, it has been suggested that SSL may prioritize general visual similarities, instead of critical subtle features in FGVR tasks, which leads to SSL's `coarse-label bias' \cite{cole2022does}. Furthermore, recent studies \cite{kim2023coreset,shu2023learning,shu2022improving} have highlighted a tendency among existing SSL methods to be distracted by task-irrelevant features, consequently failing to capture FGVR-relevant patterns.

To this end, we propose an innovative self-supervised learning strategy focusing on selectively extracting highly discriminative features while disregarding less informative, noisy ones. Our approach involves the generation of new contrastive data pairs from the latent feature space of the encoder, training the encoder to prioritize critical objects within these pairs. To facilitate this, a decoder is employed to generate data based on the latent feature space, reconstructing an image’s feature vector and its perturbed counterpart to form each data pair. As the data pairs are generated to guide the encoder to learn key features and to be invariant to variations in non-discriminative features, in each feature vector, only those dimensions associated with non-discriminative patterns are perturbed. 

Therefore, at the core of our methods are two non-discriminative feature identification techniques. Firstly, Grad-CAM \cite{selvaraju2017grad} induced from the SSL loss to the latent feature space highlights dimensions relevant to FGVR \cite{shu2022improving,shu2023learning}. Thus, we introduce greater perturbation to those less highlighted dimensions. Despite the conventional view that dimensional collapse in SSL—characterized by some encoder dimensions producing constant outputs—is undesirable \cite{chen2021exploring, li2022understanding, jing2021understanding}, recent literature \cite{cosentino2022toward, ziyin2022shapes} suggests that inducing such collapse in task-irrelevant feature dimensions yields beneficial outcomes. As shown in \cref{fig:feature-variance}, our empirical studies indicate that in encoders pre-trained by SSL methods, there are always dimensions with low variance across the dataset which cannot effectively separate data from different categories. We thus treat these low-variance dimensions as task-irrelevant and introduce perturbations to them. The two aforementioned perturbation components are then combined and applied to the latent feature vector of each image. Images are then reconstructed from the perturbed and original feature vectors to form contrastive pairs for a contrastive loss \cite{oord2018representation,chen2020simple}. Such a framework encourages the encoder to learn the key features highlighted by Grad-CAM and to reduce variance and induce collapse in those non-discriminative dimensions with low variance across the dataset in the latent space.

Our proposed fine-grained feature learning method can be incorporated into various existing SSL methods. We use SimSiam \cite{chen2021exploring} and MoCo v2 \cite{he2020momentum} as baseline methods and incorporate our proposed technique into these methods. Experiments across various fine-grained visual datasets show the effectiveness of our method. The proposed method provides a great improvement over baseline methods. Our methods built on MoCo v2 outperforms existing state-of-the-art Self-Supervised fine-grained visual recognition methods in numerous downstream tasks.

\section{Related Works}
\subsection{Self-Supervised Contrastive Learning}

Self-Supervised Learning (SSL) facilitates the learning of visual representations without the need for labeled data. Among various SSL methodologies, contrastive learning has emerged as a promising technique. With the InfoNCE loss \cite{oord2018representation} and its variants \cite{chen2020simple,he2020momentum,caron2021emerging,chen2021exploring,chuang2020debiased} being introduced as the objectives for optimization, contrastive approaches treat different views of the same image as positive data pairs, while views from different images are considered negative data pairs. The goal for the encoder is to minimize the distance between positive pairs and maximize it between negative pairs within its representation space \cite{chen2020simple,he2020momentum,wang2020understanding,caron2020unsupervised}. Methods such as BYOL \cite{grill2020bootstrap} and SimSiam \cite{chen2021exploring} rely exclusively on positive pairs. Additionally, the issue of dimensional collapse, where some encoder dimensions output constant values, is discussed in \cite{li2022understanding,jing2021understanding,chen2021exploring}, along with proposed solutions to mitigate this phenomenon. Nonetheless, recent studies \cite{ziyin2022shapes,cosentino2022toward} have shown that the collapse of dimensions associated with task-irrelevant features can enhance the performance in downstream visual recognition tasks.

\subsection{Fine-Grained Visual Recognition in Self-Supervised Learning}

While encoders pre-trained by Self-Supervised Learning (SSL) methods demonstrate transferability and generalizability in many tasks \cite{chen2020simple,he2020momentum,chen2021exploring,xiao2020should,islam2021broad}, studies \cite{cole2022does,kim2023coreset,shu2023learning} reveal SSL's limitations in capturing essential features for Fine-Grained Visual Recognition (FGVR). To enhance SSL's capability in identifying critical features, several works concentrate on refining data augmentations. Approaches such as SAGA \cite{yeh2022saga}, CAST \cite{selvaraju2021casting}, and ContrastiveCrop \cite{peng2022crafting} adopt attention-guided heatmaps to locate and better crop key objects in images. DiLo \cite{zhao2021distilling} introduces a novel augmentation by merging key image objects with different backgrounds to generate additional views. Contrary to methods that modify images directly, our approach involves perturbing feature vectors and generating realistic images from the latent feature space to enhance the encoder's discriminative capacity. Another line of research employs auxiliary neural networks connected to the encoder's convolutional layers for improving encoder's attention on salient regions. LEWEL \cite{huang2022learning} trains an additional head to adaptively aggregate features. Techniques such as CVSA \cite{di2021align} and \cite{yao2023teacher} train a network to fit segmentation annotations or outputs of pre-trained saliency detectors. Similarly, LCR \cite{shu2023learning} and SAM \cite{shu2022improving} train the network to align with Grad-CAM, treating Grad-CAM as the ground truth for the encoder's attention maps. Our method proposes training the encoder on generated data pairs to learn critical features. In addition to Grad-CAM, we use dimension variance as a criterion for identifying non-discriminative features. Low-variance dimensions where data points across the dataset are not well separated are treated less crucial. Besides, SimCore \cite{kim2023coreset} pre-trains an encoder on a target dataset, then using it to select more relevant data from a large-scaled dataset to expand the training set, upon which a new encoder is retrained for downstream tasks.

\begin{figure}[htb]
  \centering
  \includegraphics[width=1.0\textwidth]{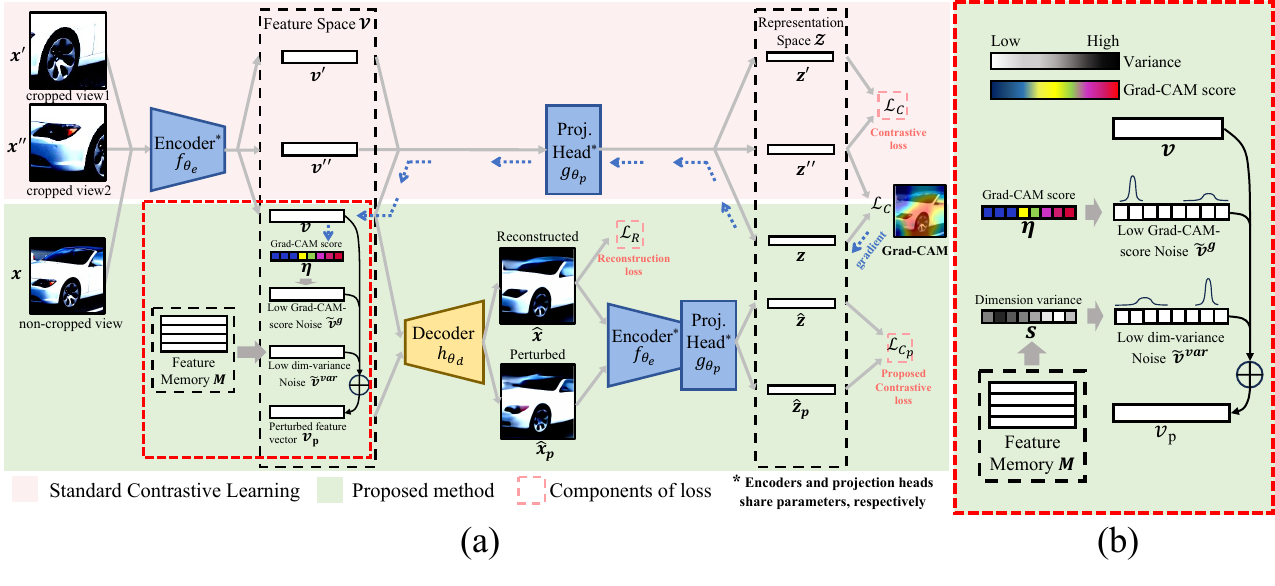}
  \caption{The overview of the proposed method. (a) Our method can be incorporated into various existing SSL methods. A decoder is utilized to generate images from both the original feature vector and its perturbed counterpart to form data pairs. The overall loss consists three terms: a conventional contrastive loss, a reconstruction loss (ensuring the decoder evolves with the encoder), and a proposed contrastive loss on the generated pairs. (b) We propose two techniques to identify and perturb non-discriminative features in a feature vector, i.e., features with low variance that fail to effectively separate data and those deemed less important by Grad-CAM induced from the SSL loss.
  }
  \label{fig:overview}
\end{figure}

\section{Method}

\subsection{Background}

\subsubsection{Self-Supervised Contrastive Learning}

Without need for labels, self-supervi- sed contrastive learning learns to represent data $\mathbf{x}\in\mathbb{R}^m$ in a lower dimensional space $\mathbb{R}^n$ by learning the similarities among data samples. Typically, in contrastive learning, the model consists two components, an encoder $f_{\theta_e}:\mathbb{R}^m\to\mathbb{R}^n$ that maps data to a latent feature space $\mathcal{V}\subseteq\mathbb{R}^n$, and a projection head $g_{\theta_p}:\mathbb{R}^n\to\mathbb{R}^k$ which projects the latent feature vectors in $\mathcal{V}\subseteq\mathbb{R}^n$ to a lower dimensional representation space $\mathcal{Z} \subseteq \mathbb{R}^k$ where the contrastive loss is applied. Formally, given a batch $\mathcal{B}$ of unlabeled data, every image $\mathbf{x}$ in it is augmented by two random augmentation $\mathcal{T}_1$ and $\mathcal{T}_2$ to acquire two views, i.e., $\mathbf{x}'=\mathcal{T}_1(\mathbf{x})$, $\mathbf{x}''=\mathcal{T}_2(\mathbf{x})$. Views augmented from the same image are considered a positive pair, while those acquired from different images form negative pairs. Two augmented views of all images form a new batch $\mathcal{B}_a$ which doubles the size of $\mathcal{B}$. The encoder and projection head are then used to represent the views in $\mathcal{Z}$, i.e., $\mathbf{z}'=g_{\theta_p}(f_{\theta_e}(\mathbf{x}'))$, $\mathbf{z}''=g_{\theta_p}(f_{\theta_e}(\mathbf{x}''))$. A contrastive loss $\mathcal{L}_C$ is then defined in $\mathcal{Z}$ space:

\begin{equation}
    \mathcal{L}_C(\mathbf{z}',\mathbf{z}'')=-\mathrm{log}\frac{\mathrm{exp}(\mathbf{z}'\cdot \mathbf{z}''/\tau)}{\sum_{\mathbf{z_i}\in \mathcal{B}_a,\mathbf{\mathbf{z}_i}\neq \mathbf{z}',\mathbf{z_i}\neq \mathbf{z}''}\mathrm{exp}(\mathbf{z'}\cdot \mathbf{z_i}/\tau)}
\end{equation}
where $\tau$ is the temperature hyperparameter. 

Although there are subtle differences between different contrastive methods, the contrastive loss are defined similarly. MoCo \cite{he2020momentum,chen2020improved} introduces a large memory of negative representations. SimSiam \cite{chen2021exploring} and BYOL \cite{grill2020bootstrap} discard negative pairs and learn solely from positive pairs.

\subsection{Overview}

The overview of our method is illustrated in \cref{fig:overview}. The essence of the proposed method is learning discriminative fine-grained visual features from synthesized data pairs which are reconstructed from latent feature vectors by a decoder $h_{\theta_d}$. During training, to ensure that the decoder follows the evolution of the encoder, we use Mean Square Error (MSE) as the loss function to optimize the decoder. Our empirical studies show that the decoder produces images of better quality when it is trained on non-augmented images. The following reconstruction loss $\mathcal{L}_R$ is calculated for each image.

\begin{equation}
    \mathcal{L}_R=\frac{1}{m}\Vert\mathbf{x}-\hat{\mathbf{x}}\Vert_2^2
\end{equation}
where $\mathbf{x}\in\mathbb{R}^m$ is non-augmented image, and $\hat{\mathbf{x}}=h_{\theta_d}(f_{\theta_e}(\mathbf{x}))$ is the reconstruction of it. 

In addition to producing $\hat{\mathbf{x}}$, we generate $\hat{\mathbf{x}}_p$ from $\mathbf{v}_p\in\mathcal{V}$, which is a perturbed version of $\mathbf{v}$ where non-discriminative features are perturbed. A positive data pair is formed between $\hat{\mathbf{x}}$ and $\hat{\mathbf{x}}_p$ from which the encoder learns discriminative features while disregarding task-irrelevant ones. In \cref{sec:gradcam-noise} and \cref{sec:low-var-noise}, we introduce two methods of identifying crucial discriminative dimensions in the latent feature space $\mathcal{V}$.

\subsection{Identifying key dimensions via Grad-CAM}
\label{sec:gradcam-noise}

Grad-CAM \cite{selvaraju2017grad}, a widely used saliency detection technique, uses the gradient of the target loss with respect to intermediate features of the network to produce an attention map highlighting regions in the features that contribute to minimizing the loss. As in our proposed self-supervised method, labels are not available during training, we thus choose the contrastive loss as the target. To identify important features within an image's feature vector $\mathbf{v}=f_{\theta_e}(\mathbf{x})\in\mathbb{R}^n$, we form positive pair between $\mathbf{x}$ and an augmented view $\mathbf{x}''$ to calculate a contrastive loss $\mathcal{L}_C(\mathbf{z},\mathbf{z}'')$, where $\mathbf{z}=g_{\theta_p}(f_{\theta_e}(\mathbf{x}))$, $\mathbf{z}''=g_{\theta_p}(f_{\theta_e}(\mathbf{x}''))$. As it is shown in \cref{fig:overview}, the Grad-CAM score vector $\bm{\eta}=\{\eta_i\}_{i=1}^n\in \mathbb{R}^n$ is calculated by gradient of the contrastive loss with respect to the feature vector $\mathbf{v}$.

\begin{equation}
    \eta_i = \mathrm{ReLU}(\frac{\partial\mathcal{L}_C(g_{\theta_p}(\mathbf{v}),g_{\theta_p}(\mathbf{v}''))}{\partial v_i}\cdot v_i)
\end{equation}
Here, $\mathbf{v}=f_{\theta_e}(\mathbf{x})$, $\mathbf{v}''=f_{\theta_e}(\mathbf{x}'')$. $v_i$ denotes the $i_{th}$ element of $\mathbf{v}$. The application of $\mathrm{ReLU}(\cdot)$ makes $\eta_i>0$ for all $i\in\{1,2,\ldots,n\}$. Note that the original Grad-CAM calculates gradient with respect to feature maps of the last convolutional layer. To measure the saliency of feature dimensions, we calculate gradient with respect to feature vectors. Higher Grad-CAM scores in $\bm{\eta}$ represent corresponding dimension's higher contribution to data discrimination in contrastive learning.

With the Grad-CAM scores, random Gaussian noise is then introduced as perturbation to $\mathbf{v}$. We first scale all elements in $\bm{\eta}$ to $[0,1]$ by min-max normalization.

\begin{equation}
    \bar{\eta}_i=\frac{\eta_i-\mathrm{min}\{\eta_j,j=1,2,\dots,n\}}{\mathrm{max}\{\eta_j,j=1,2,\dots,n\}-\mathrm{min}\{\eta_j,j=1,2,\dots,n\}}
\label{eq:min-max-norm}
\end{equation}
where $\bar{\eta}_i$ is the $i_{th}$ element of the normalized Grad-CAM score vector $\bm{\bar{\eta}}$. After the normalization, a random Gaussian noise perturbation vector $\mathbf{\Tilde{v}}^g\in \mathbb{R}^n$ is calculated as follows.

\begin{equation}
    \mathbf{\Tilde{v}}^g=\{\Tilde{v}^g_i:\Tilde{v}^g_i\sim\mathcal{N}(0,\epsilon_g\cdot(1-\bar{\eta_i}))\}_{i=1}^{n}
\label{eq:noise-gradcam}
\end{equation}
where $\epsilon_g$ is a hyperparameter that controls the standard deviation of Gaussian noise. In such a perturbation, all elements are sampled from i.i.d. zero-mean Gaussian distributions. Importantly, dimensions with lower normalized Grad-CAM scores $\bar{\eta_i}$ receive noise from a Gaussian distribution with a greater standard deviation, implying a higher likelihood of more significant noise affecting dimensions with lower Grad-CAM scores. And on average, key dimensions with higher Grad-CAM scores are affected less which helps preserve crucial features in the original images.

\begin{figure}[h]
  \centering
  \includegraphics[width=0.97\textwidth]{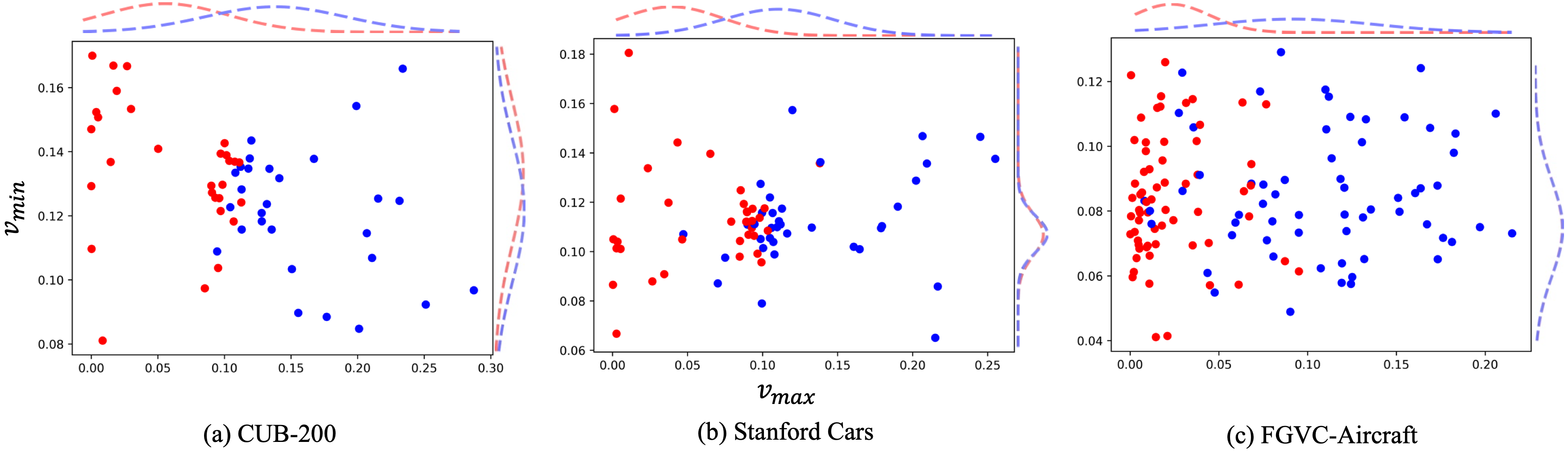}
  \caption{An illustration of data distribution in the feature space of encoders pre-trained by MoCo v2 \cite{chen2020improved}. Blue and red dots represent feature vectors of two categories' data from 3 fine-grained datasets, CUB-200 \cite{wah2011caltech}, Stanford Cars \cite{krause20133d}, and FGVC-Aircraft \cite{maji2013fine}. $v_{min}$ and $v_{max}$ are the dimensions in the feature space where data has the minimal and maximal variance across the dataset. Probability density curve fitting of each category along each dimension is attached to the corresponding axis. Different classes are separated much better along $v_{max}$ than $v_{min}$.
  }
  \label{fig:feature-variance}
\end{figure}

\subsection{Determining feature's task-relevance via dimension variance}
\label{sec:low-var-noise}

In addition to the feature perturbation technique elaborated in \cref{sec:gradcam-noise}, we propose another technique to determine and perturb task-irrelevant dimensions. 

In SSL, dimensional collapse is a phenomenon where some encoder dimensions output constant values \cite{li2022understanding,jing2021understanding,chen2021exploring,hua2021feature}. As data points are not separated along such dimensions, these dimensions can not be used to perform downstream visual recognition. However, recent works \cite{cosentino2022toward,ziyin2022shapes} suggest that collapse of dimensions which are related to downstream task-irrelevant features can be beneficial. In spite of the potential benefit, how to induce beneficial dimensional collapse is not illustrated by existing SSL studies. 

As it is illustrated in \cref{fig:feature-variance}, our empirical studies show that, in the latent feature space of encoders pre-trained by SSL methods, typically, data points are not well separated along dimensions with low variance. Variance along these dimensions thus introduces noise to downstream classification. Therefore, we treat such dimensions as task-irrelevant and propose a technique to induce collapse in these dimensions to guide the encoder's to be invariant to variations of such features. This technique start with estimating the dataset's variance along each feature dimension in a feature vector memory bank $\mathbf{M}\in \mathbb{R}^{D\times n}$ of size $D$. During training, whenever the encoder is provided with a batch of data, its feature vectors will be stored in the memory to replace the oldest batch of feature vectors in it. Variance of each dimension across the dataset can be approximated in $\mathbf{M}$ to get a variance vector $\mathbf{s}=\{s_i:\sigma^2(\mathbf{\bar{w}}_i),i=1,2,\dots,n\}\in\mathbb{R}^n$. Here $\mathbf{\bar{w}}_i\in \mathbb{R}^D$ is the $\ell_2$-normalized $i_{th}$ column vector of $\mathbf{M}$. A feature represented by dimension $i$ is considered less discriminative if its corresponding variance $s_i<\kappa$ where $\kappa$ is a threshold hyperparameter. To introduce random noise to those less discriminative dimensions, similar to \cref{eq:min-max-norm}, we first apply min-max normalization to $\mathbf{s}$ to acquire $\mathbf{\bar{s}}=\{\bar{s_i}\}_{i=1}^n$. We then calculate the random noise vector $\mathbf{\Tilde{v}}^{var}=\{\Tilde{v}^{var}_i\}_{i=1}^{n}$ to be applied.

\begin{equation}
    \Tilde{v}^{var}_i=\left\{
    \begin{aligned}
        &u_i \sim \mathcal{N}(0,\epsilon_{var}\cdot(1-\bar{s_i})) &&\mathrm{if}\ s_i<\kappa  \\
        &0  &&\mathrm{otherwise}
    \end{aligned}\right.
\label{eq:noise-low-var}
\end{equation}
Here, $\epsilon_{var}$ defines the standard deviations of the i.i.d. Gaussian distributions. Similar to $\mathbf{\Tilde{v}}^g$, $\mathbf{\Tilde{v}}^{var}$ introduces greater noise to lower-variance dimensions. 

\subsection{Learning from reconstructed data pairs}

With the two feature perturbation techniques proposed in \cref{sec:gradcam-noise} and \cref{sec:low-var-noise}, we can finally perturb the feature vector $\mathbf{v}$ by adding the two random Gaussian noise to it to obtain its perturbed version $\mathbf{v}_p$, i.e., $\mathbf{v}_p=\mathbf{v}+\mathbf{\Tilde{v}}^g+\mathbf{\Tilde{v}}^{var}$. As it is shown in \cref{fig:overview}, $\mathbf{v}$ and $\mathbf{v}_p$ are reconstructed by the decoder to produce $\mathbf{\hat{x}}$ and $\mathbf{\hat{x}}_p$, respectively. In $\mathbf{\hat{x}}_p$, key patterns from $\mathbf{\hat{x}}$ are preserved, while those contribute less to Grad-CAM or deemed less discriminative across the dataset by the low-variance criterion are perturbed. 

We then form a positive pair between $\mathbf{\hat{x}}$ and $\mathbf{\hat{x}}_p$, addressing extracting key features and disregarding non-discriminative ones. To this end, we propose to pass the representations $\mathbf{\hat{z}}=g_{\theta_p}(f_{\theta_e}(\mathbf{\hat{x}}))$ and $\mathbf{\hat{z}}_p=g_{\theta_p}(f_{\theta_e}(\mathbf{\hat{x}}_p))$ to a contrastive loss $\mathcal{L}_{C_p}$.

\begin{equation}
    \mathcal{L}_{C_p}(\mathbf{\hat{z}},\mathbf{\hat{z}}_p)=-\mathrm{log}\frac{\mathrm{exp}(\mathbf{\hat{z}}\cdot \mathbf{\hat{z}}_p/\tau)}{\sum_{\mathbf{\hat{z}_i}\in \mathcal{B}_r,\mathbf{\mathbf{\hat{z}}_i}\neq \mathbf{\hat{z}},\mathbf{\hat{z}_i}\neq \mathbf{\hat{z}}_p}\mathrm{exp}(\mathbf{\hat{z}}\cdot \mathbf{\hat{z}_i}/\tau)}
\label{eq:perturbation-loss}
\end{equation}
where $\mathcal{B}_r$ is the set of all reconstructed and perturbed images from the current batch $\mathcal{B}$. By forming a positive pair between $\mathbf{\hat{x}}$ and $\mathbf{\hat{x}}_p$, \cref{eq:perturbation-loss} requires the encoder to be invariant to those perturbed less discriminative features.

Finally, we write the training loss $\mathcal{L}$ of our method as follows.

\begin{equation}
    \mathcal{L}=\mathcal{L}_C+\alpha\cdot\mathcal{L}_R+\nu\cdot\mathcal{L}_{C_p}
\label{eq:overall-loss}
\end{equation}

$\alpha$ and $\nu$ are hyperparameters that control the weight of $\mathcal{L}_R$ and $\mathcal{L}_{C_p}$ in training. The pseudocode of our method is provided in Appendices.

\begin{table}[htb]

  \caption{Performance comparison on three datasets. Our method is compared with MoCo v2 \cite{chen2020improved} and ResNet50 supervised pre-trained on ImageNet-1k \cite{imagenet_cvpr09}. Top-1 classification accuracy (in\%) is reported when model is evaluated on 100\%, 50\%, and 20\% of all labels. Rank-1, rank-5, and mAP (in\%) in image retrieval are reported.
  }
  \label{tab:ours-vs-moco}
  \centering
  \scalebox{0.8}{
  \setlength{\tabcolsep}{3.0mm}{
  \begin{tabular}{l|l|ccc|ccc}
    \toprule
    & & \multicolumn{3}{c}{Classification} & \multicolumn{3}{c}{Image Retrieval}\\
    \cmidrule(lr){3-5}
    \cmidrule(lr){6-8}
    
    Dataset & Methods & 100\% & 50\% & 20\% & rank-1 & rank-5 & mAP \\
    \midrule
            & ResNet50 & 63.06      & 55.71      & 42.24      & 40.39      & 68.94      & 15.88\\
    CUB-200 & MoCo V2  & 63.98      & 56.35      & 42.63      & 39.72      & 67.14      & 15.91 \\
            & Ours     & \bf{66.17} & \bf{60.84} & \bf{49.69} & \bf{42.06} & \bf{69.59} & \bf{19.70} \\
    \midrule
                  & ResNet50 & 61.41      & 49.85      & 33.57      & 29.28      & 54.66      &7.01\\
    Stanford Cars  & MoCo V2  & 62.02      & 51.08      & 35.44      & 30.51      & 56.15      & 7.13\\
                  & Ours     & \bf{65.60} & \bf{54.36} & \bf{40.24} & \bf{35.81} & \bf{61.94} & \bf{10.02}\\
    \midrule
                  & ResNet50 & 49.79      & 41.02      & 33.61      & 29.48      & 51.39      &10.17\\
    FGVC-Aircraft & MoCo V2  & 51.13      & 44.34      & 36.42      & 30.02      & 52.87      & 11.24\\
                  & Ours     & \bf{55.28} & \bf{49.37} & \bf{41.10} & \bf{33.27} & \bf{56.80} & \bf{12.69}\\
  \bottomrule
\end{tabular}}}
\end{table}

\section{Experiments}
\label{sec:experiments}

\subsection{Experiment Settings}
\subsubsection{Datasets.} 
Experiments are conducted across five fine-grained visual datasets. We adopt on three widely used fine-grained datasets. \texttt{Caltech-UCSD Birds 200 -2011 (CUB-200)} dataset \cite{wah2011caltech} contains 5994 training data and 5794 testing data of 200 categories of birds. \texttt{Stanford Cars (Cars)} \cite{krause20133d} has 196 classes of car models where 8144 data and 8041 data are in its training and testing split respectively. \texttt{FGVC-Aircraft (Aircraft)} \cite{maji2013fine} has 100 classes where 6667 images are for training and 3333 images are for testing. Additionally, we consider \texttt{German Traffic Sign Recognition Benchmark (GTSRB)} \cite{Houben-IJCNN-2013} which contains 43 classes of traffic signs. This dataset is usually used in autonomous driving and smart cities development. We take 4800 images for training and 3750 images for testing from \texttt{GTSRB}. We also evaluate the effectiveness of our method on \texttt{ISIC2017} \cite{codella2018skin}, a medical image dataset with 3 categories of skin lesion analysis where 2000 images are for training and 600 images are for testing.

\subsubsection{Training Settings.} All methods adopt ResNet-50 \cite{he2016deep} as the encoder backbone where weights are initialized by loading ImageNet-1k \cite{ILSVRC15} pre-trained model. Using two state-of-the-art SSL method, MoCo v2 \cite{chen2020improved} and SimSiam \cite{chen2021exploring}, as the baseline methods, we incorporate our proposed method in their framework. For the sake of fair comparison, encoders of all methods are pre-trained for 100 epochs. And the training batch size is set to 128 for all methods. More detailed encoder pre-training settings are provided in Appendices.

Additionally, in our method, we use a feature vector memory bank of size $D=5632$. For the training loss in \cref{eq:overall-loss}, we choose $\alpha=1$ and $\nu=0.5$. When introducing noise described in \cref{eq:noise-gradcam} and \cref{eq:noise-low-var}, we choose $\epsilon_g=0.1$, $\epsilon_{var}=0.05$, and $\kappa=0.02$. To ensure the quality of reconstructed images, before encoder training, we freeze the encoder parameters and pre-train the decoder on target datasets by a reconstruction loss. We provide decoder pre-training details in Appendices.

\subsubsection{Performance Evaluation Protocols.} 

Linear evaluation is a widely adopted protocol for assessing the performance of learned representations in visual recognition. This approach freezes the parameters of the pre-trained encoder and attaches a linear classifier to it. The classifier is then trained to perform classification. Our linear evaluation setup follows \cite{he2020momentum}, detailed further in the Appendices.

The task of image retrieval \cite{jang2021self,shu2023learning,xiao2020should} serves as another pivotal method for evaluating the performance of representation learning. Without adjusting any model parameters, it searches the nearest neighbors of a query image in the latent feature space for images that share the same label with the query image. The effectiveness of the evaluated model is quantified by recording the proportion of retrieved images that fall into the same categories as the query image. We present three commonly utilized metrics of retrieval performance: rank-1, rank-5, and mean Average Precision (mAP).

Furthermore, to illustrate the effect of the proposed method, attention maps generated by Grad-CAM on encoders trained by different methods are compared. Images from the training datasets, along with their corresponding reconstructed and perturbed versions, are also provided.

\begin{figure}[h]
  \centering
  \includegraphics[width=0.76\textwidth]{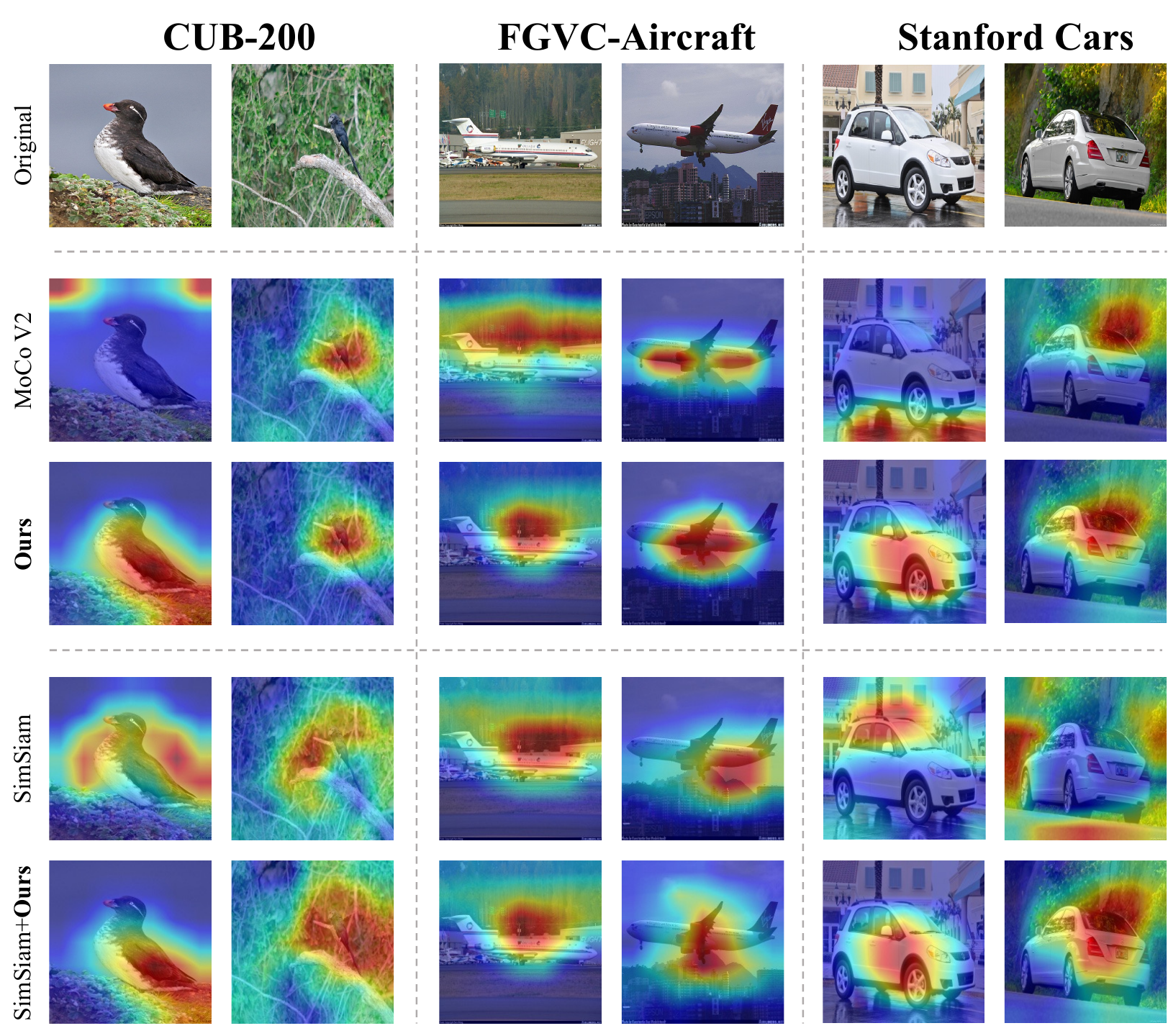}
  \caption{Grad-CAM attention visualized on images. Our proposed method is incorporated into MoCo v2 and SimSiam and compared with them.
  }
  \label{fig:gradcam}
\end{figure}

\subsection{Main Results}

\subsubsection{Comparison between the proposed method and baseline methods}

As among all configurations of our method, the one based on MoCo v2 achieves the best overall performance, this configuration is thus called \textbf{`ours'} in the experiments. We first compare our method with MoCo v2 in \cref{tab:ours-vs-moco}. `ResNet50' in \cref{tab:ours-vs-moco} represents the encoder supervised pre-trained on ImageNet-1k \cite{imagenet_cvpr09}.

In linear evaluation, we evaluate pre-trained encoders using different proportions of labels. Across all three datasets, our proposed method achieves an average top-1 accuracy improvement of 3.31\%, 4.27\%, and 5.51\% over MoCo v2, with 100\%, 50\%, and 20\% of all labels, respectively. This advancement highlights the efficacy of our data pairs generation technique, enabling great performance in classification tasks. Notably, the remarkable improvement in label-insufficient scenarios highlights our method's ability to learn more generalizable features from unlabeled data.

We further evaluate the effectiveness of our method by image retrieval tasks. Performance in these tasks serves as a measure of semantic consistency in the learned latent feature space. Our method outperforms MoCo v2 on all datasets in rank-1, rank-5, and mAP. Invariant to irrelevant patterns, our approach ensures that images from the same category, which exhibit discriminative patterns, are more closely clustered in the feature space. This distribution enhances performance in image retrieval.

Additionally, we visualize Grad-CAM attention in MoCo v2 and our method in \cref{fig:gradcam}. Unlike MoCo v2, which may concentrate on background regions irrelevant to visual recognition, our method exhibits enhanced precision in identifying and focusing on pivotal objects within the images, effectively minimizing the influence of background distractions.

Furthermore, we also integrate the proposed technique into a state-of-the-art negative pair-free method SimSiam \cite{chen2021exploring}, denoted as `SimSiam+ours' in our experiments. Our framework significantly outperforms the original SimSiam in both linear evaluation and image retrieval tasks, as demonstrated in \cref{tab:sota-ssl-fgvr}. Attention comparison is also visualized by Grad-CAM and compared with SimSiam in \cref{fig:gradcam}.

\begin{table}[tb]
  \caption{Performance comparison on ISIC2017 \cite{codella2018skin} and GTSRB\cite{Houben-IJCNN-2013}. Top-1 classification accuracy (in \%) and rank-1 (in \%) of image retrieval is reported.
  }
  \label{tab:more-datasets}
  \centering
  \scalebox{0.8}{
  \setlength{\tabcolsep}{3.0mm}{
  \begin{tabular}{l|cc|cc}
    \toprule
    &  \multicolumn{2}{c}{ISIC2017} & \multicolumn{2}{c}{GTSRB}\\
    \cmidrule(lr){2-3}
    \cmidrule(lr){4-5}
    
    Methods & Classification & Retrieval & Classification & Retrieval \\
    \midrule
    MoCo V2 \cite{chen2020improved}        & 64.33      & 66.34      & 88.25      & 97.46      \\
    Ours  & \bf{66.92} & \bf{67.90} & \bf{91.67} & \bf{98.59} \\
    
    \midrule
    SimSiam \cite{chen2021exploring}       & 58.83      &  64.46     & 81.10      & 92.81      \\
    SimSiam + Ours & \bf{63.78} & \bf{65.55} & \bf{86.94} & \bf{94.42} \\
  \bottomrule
\end{tabular}}}
\end{table}

To comprehensively assess the effectiveness of our proposed method, we utilize two more fine-grained datasets: the traffic sign visual dataset \texttt{GTSRB} and the medical image dataset \texttt{ISIC2017}. The results, as shown in \cref{tab:more-datasets}, indicate that our approach enhances the performance of Self-Supervised Learning (SSL), demonstrating its potential in real-world applications for FGVR tasks.

\begin{table}[htb]
  \caption{Comparison with state-of-the-art self-supervised FGVC methods. Supervised training is also included. Top-1 classification accuracy (in \%) and rank-1 image retrieval (in \%) are reported.
  }
  \label{tab:sota-ssl-fgvr}
  \centering
  \scalebox{0.82}{
  \setlength{\tabcolsep}{3.0mm}{
  \begin{tabular}{l|ccc|ccc}
    \toprule
    & \multicolumn{3}{c}{Classification} & \multicolumn{3}{c}{Image Retrieval} \\
    \cmidrule(lr){2-4} 
    \cmidrule(lr){5-7}

    Method & CUB-200 & Cars & Aircraft & CUB-200 & Cars & Aircraft \\
    \midrule
    supervised                              & 77.46   & 88.60   & 85.93   & - & - & - \\
    \midrule
    Dino \cite{caron2021emerging}           & 16.74   & 14.33   & 12.07   & -   & - &-      \\
    SimSiam \cite{chen2021exploring}        & 46.75   & 45.72   & 38.52   & 16.24   & 12.45  & 18.49    \\
    MoCo V2 \cite{chen2020improved}         & 63.98   & 62.02   & 51.13   & 39.72   & 30.51   & 30.02   \\
    DiLo \cite{zhao2021distilling}          & 62.97   & -   & -   & -   & - &-       \\
    CVSA \cite{di2021align}                 & 63.02   & -   & -   & -   & - &-         \\
    LEWEL \cite{huang2022learning}          & 64.59   & 62.91   & 51.90   & 39.91   & 32.36  &  31.09  \\
    ContrastiveCrop \cite{peng2022crafting} & 64.23   & 63.29   & 52.04   & 39.84   & 32.71  &  30.37  \\
    SAM-SSL-Bilinear \cite{shu2022improving}& 64.94   & 62.85   & 52.83   & 40.08   & 33.19  & 30.52   \\
    LCR \cite{shu2023learning}              & 65.24   & 63.96   & 53.22   & 41.26   & 34.74  & 31.55\\
    
    \midrule
    SimSiam+Ours  & 57.80      & 53.63      & 47.50      & 24.67      &  19.72     &  24.56\\
    Ours          & \bf{66.17} & \bf{65.60} & \bf{55.28} & \bf{42.06} & \bf{35.81}  & \bf{33.27} \\
  \bottomrule
\end{tabular}}}
\end{table}

\subsubsection{Comparison with state-of-the-art self-supervised FGVR methods}

In this section, we compare our proposed method against existing self-supervised learning (SSL) techniques renowned for their enhanced fine-grained visual recognition capabilities\cite{zhao2021distilling,di2021align,huang2022learning,peng2022crafting,shu2023learning,shu2022improving}, as detailed in \cref{tab:sota-ssl-fgvr}. Methods like Dino \cite{caron2021emerging}, SimSiam\cite{chen2021exploring}, and MoCo v2\cite{he2020momentum}, which are not optimized for fine-grained feature extraction are also listed. Supervised training performance is included to provide a comprehensive comparison. Top-1 accuracy in linear evaluation and rank-1 in image retrieval tasks are reported.

In \cref{tab:sota-ssl-fgvr}, our proposed method achieves the best overall performance in both image retrieval and linear evaluation tasks. DiLo \cite{zhao2021distilling}, CVSA \cite{di2021align}, and ContrastiveCrop \cite{peng2022crafting} innovate with novel data augmentation techniques which directly modify the original images. In contrary, our method generates more realistic images from the learned feature space, highlighting the learning of FGVR-related features. And unlike SAM \cite{shu2022improving} and LCR \cite{shu2023learning} which train an auxiliary network to directly fit the encoder's attention to Grad-CAM, our method learns from generated data to highlight dimensions with high Grad-CAM scores and introduce dimensional collapse to non-discriminative features.

\begin{figure}[H]
  \centering
  \includegraphics[height=10.2cm]{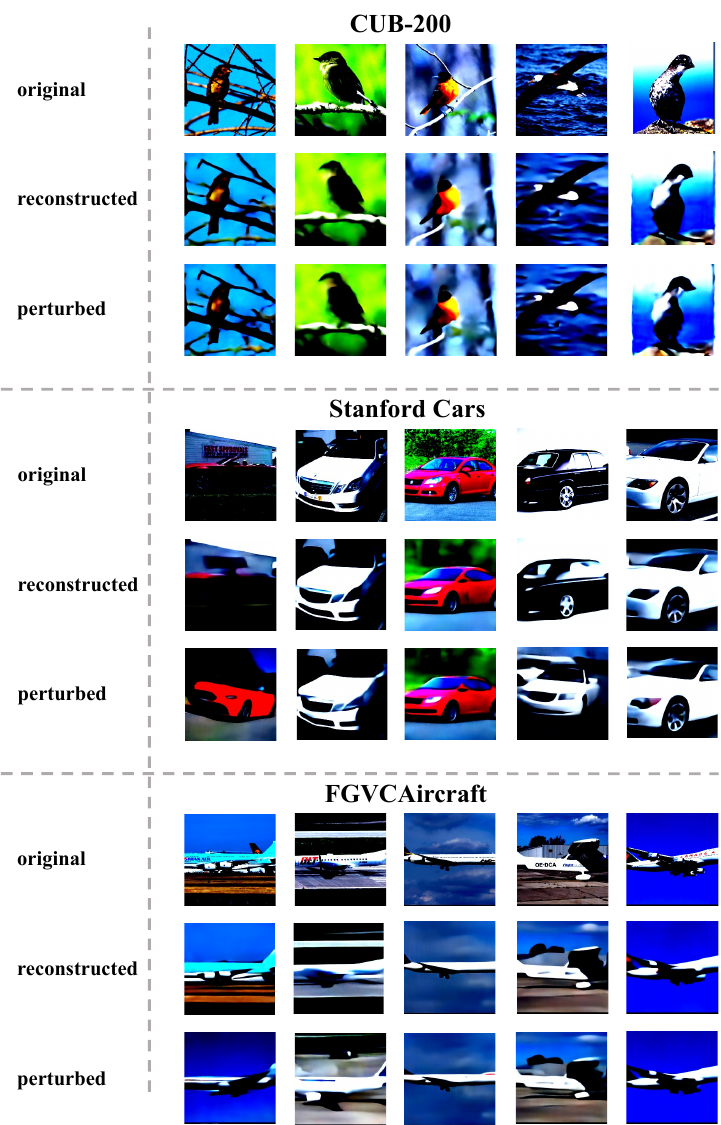}
  \caption{Generated data pairs on CUB-200, Stanford Cars, and FGVC-Aircraft. The original images are also included.
  }
  \label{fig:images}
\end{figure}

\subsubsection{Generated data pairs of the proposed method}

To better understand why the proposed technique enhances the fine-grained visual feature extraction capability, we show the generated data pairs from different datasets in \cref{fig:images}. The perturbed images, as illustrated in \cref{fig:images}, show that they largely retain the original data's key objects, with modifications primarily appearing as subtle changes in background or less important regions, e.g., changes of the tree's branch behind a bird, alterations in vehicle light's textures and adjustments in an airplane's exterior finish. These modifications do not affect the defining features of the subjects. Remarkably, some perturbations lead to entirely new objects that maintain the original's identity. For instance, transforming a side view of a car into a front view, or depicting an aircraft in flight from a grounded position. These images, obtained by modifying latent semantics of original images, are difficult to obtain through traditional data augmentation techniques defined in the original data space. They efficiently guide the encoder in identifying which features to prioritize and which to ignore, enhancing its learning performance in FGVR.

\begin{figure}[h]
  \centering
  \includegraphics[height=5.5cm]{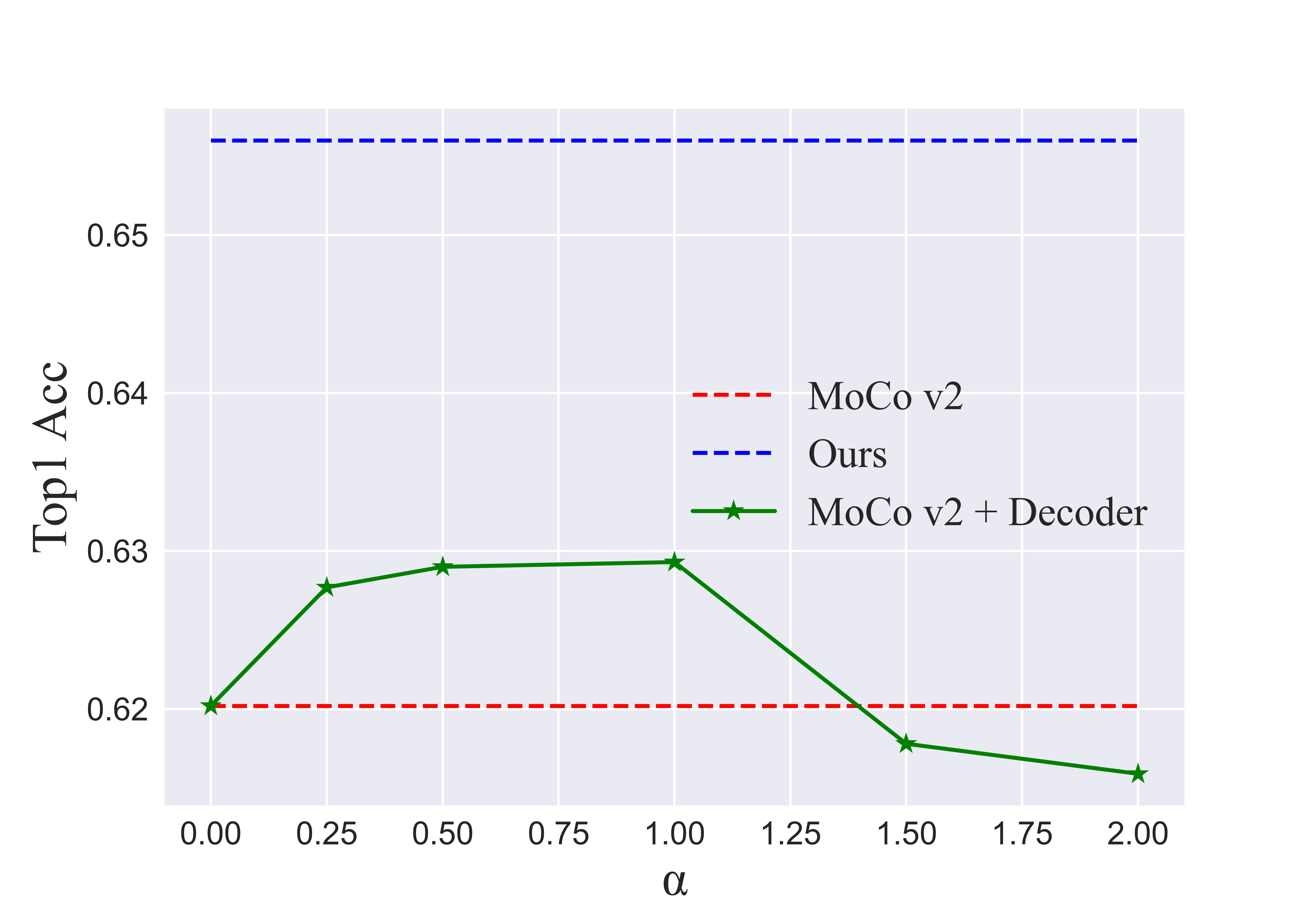}
  \caption{Performance comparison of encoders trained by $\mathcal{L}_C+\alpha\cdot\mathcal{L}_R$ with respect to different $\alpha$ value (green solid line). Top-1 classification accuracy on Stanford Cars is reported. MoCo v2 (red dashed line) and our method (blue dashed line) are included for comparison.
  }
  \label{fig:ablation}
\end{figure}

\subsection{Effectiveness of the reconstruction loss in contrastive learning}

As described in \cref{eq:overall-loss}, our model's overall training loss, $\mathcal{L}$, includes a reconstruction loss term, $\mathcal{L}_R$. To assess $\mathcal{L}_R$'s effect on self-supervised contrastive learning, we incorporate a decoder into MoCo v2 and train the encoder by a loss $\mathcal{L}_C+\alpha\cdot\mathcal{L}_R$ on the Stanford Cars dataset, varying $\alpha$ in the loss function. The results, shown in \cref{fig:ablation}, indicate that $\alpha$ values of 0.5 and 1.0 enhance top-1 classification accuracy the most over MoCo v2. Generally, $\mathcal{L}_R$ provides a modest improvement (less than 1\%) to Self-Supervised FGVC.

\subsection{Effectiveness of the two feature perturbation techniques}

As detailed in \cref{sec:gradcam-noise} and \cref{sec:low-var-noise}, two non-discriminative feature perturbation techniques are proposed to introduce noise $\mathbf{\Tilde{v}}^g$ and $\mathbf{\Tilde{v}}^{var}$, respectively. We conduct further experiments to assess the effectiveness of each technique in identifying and perturbing task-irrelevant features. In \cref{tab:ablation-two-noise}, we evaluate three different configurations of our method: (1) Ours: $\mathbf{v}_p$ is obtained by adding both noise components to $\mathbf{v}$, i.e., $\mathbf{v}_p=\mathbf{v}+\mathbf{\Tilde{v}}^g+\mathbf{\Tilde{v}}^{var}$; (2) Ours - Grad-CAM: $\mathbf{v}_p$ is obtained by adding only the noise generated by the low Grad-CAM scores criterion, i.e., $\mathbf{v}_p=\mathbf{v}+\mathbf{\Tilde{v}}^g$; (3) Ours - low-var: $\mathbf{v}_p=\mathbf{v}+\mathbf{\Tilde{v}}^{var}$.

As shown in \cref{tab:ablation-two-noise}, all configurations achieve competitive results comparing with existing state-of-the-art self-supervised FGVR methods. And when combining two noise components $\mathbf{\Tilde{v}}^g$ and $\mathbf{\Tilde{v}}^{var}$, our methods achieves the best overall performance.

\begin{table}[htb]
  \caption{Comparison with state-of-the-art self-supervised FGVC methods. Supervised training is also included. Top-1 classification accuracy (in \%) and rank-1 image retrieval (in \%) are reported.
  }
  \label{tab:ablation-two-noise}
  \centering
  \scalebox{0.9}{
  \setlength{\tabcolsep}{2.3mm}{
  \begin{tabular}{l|ccc|ccc}
    \toprule
    & \multicolumn{3}{c}{Classification} & \multicolumn{3}{c}{Image Retrieval} \\
    \cmidrule(lr){2-4} 
    \cmidrule(lr){5-7}

    Method & CUB-200 & Cars & Aircraft & CUB-200 & Cars & Aircraft \\
    \midrule
    MoCo V2 \cite{chen2020improved}         & 63.98   & 62.02   & 51.13   & 39.72   & 30.51   & 30.02   \\
    Ours - Grad-CAM       & 66.04      & 64.18      & 54.19      & 41.02      & 35.08      & 32.14 \\
    Ours - low-var       & 65.91      & 64.34      & 53.11      & 41.69      & 34.55      & 31.23 \\
    Ours          & \bf{66.17} & \bf{65.60} & \bf{55.28} & \bf{42.06} & \bf{35.81}  & \bf{33.27} \\
  \bottomrule
\end{tabular}}}
\end{table}

\section{Conclusion}

To enhance the performance of Self-Supervised Learning (SSL) in Fine-grained Visual Recognition (FGVR) tasks, this paper introduces a novel approach where an encoder learns discriminative features from generated images. By introducing noise to features deemed non-discriminative by two proposed criteria, we generate synthetic data from both the original and perturbed feature vectors by a decoder, thus forming data pairs that emphasize learning key features for FGVR. Our approach outperforms existing methods across various datasets in many downstream tasks. While the proposed approach also offers a modest boost to SSL performance in non-fine-grained visual recognition tasks—as detailed in the Appendices—the gains are notably more substantial in FGVR contexts. The refinement of our methodology for application to large-scaled general datasets remains an avenue for future research works.

\noindent \textbf{Acknowledgment.}
This material is based upon work supported by the National Science Foundation under Grant No. 1956313.
%
%

\bibliographystyle{splncs04}
\bibliography{egbib}
\newpage
\appendix
\begin{center}
    \Large{\bf Appendix}
\end{center}
\section{Pseudo-code}

As shown in \cref{alg:pseudo-code}, we conclude algorithm flow of the proposed method and present it as Pytorch-like style pseudo-code.

\begin{breakablealgorithm}
\caption{Pytorch-like style pseudo-code of our method built on MoCo v2.}
\label{alg:pseudo-code}
\begin{algorithmic}
 \Require{Initial query and key encoder parameters $\theta_q$ and $\theta_k$, initial query and key projection head parameters $\theta_{p_q}$ and $\theta_{p_k}$, initial decoder parameters $\theta_d$, and unlabeled dataloader. MoCo's hyperparameters: momentum $m$, temperature $\tau$. Our method's hyperparameters: $\alpha, \nu, \epsilon_g, \epsilon_{var}, \kappa$} \\
\For{x in \rm dataloader}

\State \textcolor{ForestGreen}{"""--------- Calculate $\mathcal{L}_C$ ---------"""} 
\State \textcolor{ForestGreen}{\# Acquire two views by stochastic augmentations} 
\State $\rm x' = T(x)$ 
\State $\rm x'' = T(x)$ 
\State \textcolor{ForestGreen}{\# Update key model} 
\State $\rm \theta_k = m*\theta_k + (1-m)*\theta_q$ 
\State $\rm \theta_{p_k} = m*\theta_{p_k} + (1-m)*\theta_{p_q}$ 
\State \textcolor{ForestGreen}{\# Calculate key and query representations} 
\State $\rm q=proj\_head\_q(encoder\_q(x'))$ 
\State $\rm k=proj\_head\_k(encoder\_k(x'')).detach()$ 
\State \textcolor{ForestGreen}{\# Representation memory enqueue \& dequeue} 
\State$\rm enqueue\_dequeue(Queue, k)$ 
\State\textcolor{ForestGreen}{\# Calculate positive \& negative pair logits} 
\State$\rm l\_pos=einsum('ck,ck->c', [q,k])$ 
\State$\rm l\_neg=einsum('ck,kj->cj', [q,Queue])$ 
\State\textcolor{ForestGreen}{\# Calculate contrastive loss $\mathcal{L}_C$} 
\State$\rm logits=cat([l\_pos,l\_neg], dim=1) / \tau$ 
\State$\rm target=zeros(len(l\_pos))$ 
\State$\rm loss\_C=cross\_entropy(logits, target)$ 

\State\textcolor{ForestGreen}{"""--------- Calculate $\mathcal{L}_R$ ---------"""} 
\State$\rm \hat{x}=decoder(encoder\_q(x))$ 
\State$\rm loss\_R=MSE\_Loss(x,\hat{x})$ 

\State\textcolor{ForestGreen}{"""--------- Calculate $\mathcal{L}_{C_p}$ ---------"""} 
\State$\rm v=encoder\_q(x)$ 
\State\textcolor{ForestGreen}{\# Feature memory bank enqueue \& dequeue} 
\State$\rm enqueue\_dequeue(M, v)$ 
\State$\rm z=proj\_head\_q(v)$ 
\State$\rm z''=proj\_head\_q(encoder\_q(x''))$ 
\State\textcolor{ForestGreen}{\# Contrastive\_loss($\cdot$,$\cdot$) returns the SimCLR loss when given two views}
\State$\rm l\_c=contrastive\_loss(z,z'')$ 
\State$\rm \eta=get\_gradcam(l\_c,v)$ \textcolor{ForestGreen}{\# size: n}
\State$\rm inverse\_normed\_\eta = 1 - (\eta - min(\eta))/(max(\eta)-min(\eta))$ \textcolor{ForestGreen}{\# size: n} 
\State\textcolor{ForestGreen}{\# Calculate random noise derived from Grad-CAM scores} 
\State$\rm \Tilde{v}_{var}=normal(mean=0, std=inverse\_normed\_\eta)$ \textcolor{ForestGreen}{\# size: n} 
\State\textcolor{ForestGreen}{\# $\ell_2$-normalize each column in $M$ and calculate variance in each column}
\State$\rm \bar{s}=var(normalize(M, dim=0), dim=0) $
\State$\rm high\_var\_mask=\bar{s}<\kappa $
\State$\rm inverse\_normed\_\bar{s}=1 - (\bar{s} - min(\bar{s}))/(max(\bar{s})-min(\bar{s}))$
\State\textcolor{ForestGreen}{\# Calculate random noise derived from variance and mask out} 
\State\textcolor{ForestGreen}{\# high-variance dimensions} 
\State $\rm \Tilde{v}_{g}=normal(mean=0, std=inverse\_normed\_\bar{s})$
\State $\rm \Tilde{v}_{g}=\Tilde{v}_{g}[high\_var\_mask]$

\State $\rm v_p=v+\Tilde{v}_{g}+\Tilde{v}_{var}$
\State $\rm \hat{x}_p=decoder(v_p)$  \textcolor{ForestGreen}{\# Reconstruct the perturbed feature vector} 
\State $\rm \hat{z}=proj\_head\_q(encoder\_q(\hat{x}.detach()))$
\State $\rm \hat{z}_p=proj\_head\_q(encoder\_q(\hat{x}_p.detach()))$
\State $\rm reps=cat([\hat{z}, \hat{z}_p],0)$
\State\textcolor{ForestGreen}{\# Contrastive loss in generated data pairs}
\State $\rm loss\_Cp=simclr\_loss(reps)$  
\State $\rm loss=loss\_C+loss\_R+loss\_Cp$
\State $\rm loss.backward()$
\State\textcolor{ForestGreen}{\# Update query encoder, projection head, and decoder parameters}
\State $\rm update(\theta_q)$
\State $\rm update(\theta_{p_q})$
\State $\rm update(\theta_d)$
\EndFor \\
\Ensure{Query encoder parameters $\theta_q$.}
\end{algorithmic}
\end{breakablealgorithm}

\section{Experiment Details}
\subsection{Encoder Pre-training}
For MoCo v2 \cite{chen2020improved} and our MoCo v2 based implementation, we follow the settings of \cite{chen2020improved}. Specifically, in pre-training, we adopt a cosine scheduled SGD with a initial learning rate of 0.03, a weight decay of $10^{-4}$, and a momentum of 0.9. The momentum for key encoder update is 0.999. Size of the representation memory bank is set to 65536.

For SimSiam \cite{chen2021exploring} and our SimSiam based implementation, we use the settings of \cite{chen2021exploring}. We use an SGD optimizer with an initial learning rate of 0.05, a weight decay of $10^{-4}$, and a momentum of 0.9. The learning rate for the encoder is cosine scheduled while the predictor is trained with a constant learning rate.

The random data augmentation for training data is composed of \texttt{Random- ResizedCrop}, \texttt{RandomHorizontalFlip}, \texttt{ColorJitter}, \texttt{RandomGreyScale}, and \texttt{GaussianBlur}.

\subsection{Decoder Initialization}

To enhance the quality of the generated images, we pre-train a decoder before training the encoder. This decoder is connected to the encoder, whose parameters remain fixed during the training phase, and it is optimized using a Mean Squared Error (MSE) reconstruction loss. Training is conducted using an Adam optimizer, with an initial learning rate of $10^{-4}$. The decoder is trained for 200 epochs with a batch size of 128. During training, we solely apply a \texttt{Resize} transformation to adjust the data to a size of $224\times224$.

\subsection{Linear Evaluation Protocol}
For all linear evaluation experiments, we train a linear classifier connected to the frozen pre-trained encoder on labeled data. Following \cite{he2020momentum}, we adopt an SGD optimizer with an initial learning rate of 30.0, a momentum of 0.9, and a weight decay of 0. Each linear classifier is trained for 100 epochs with a batch size of 128.

All training data is applied \texttt{RandomResizedCrop} of size $224\times224$ and \texttt{Random- HorizontalFlip}. Testing data is applied a \texttt{CenterCrop} of size $224\times224$.

\section{Experiments on Non-fine-grained Datasets}

To further evaluate the proposed method's effectiveness, we conduct experiments on non-fine-grained visual datasets, \texttt{CIFAR-100} \cite{krizhevsky2009learning} and \texttt{ImageNet-100} \cite{imagenet_cvpr09,tian2020contrastive}. \texttt{ImageNet-100} is a 100-class subset of \texttt{ImageNet-1k}. We follow \cite{tian2020contrastive} to sample the 100 classes. As shown in \cref{tab:non-fine-grained}, benefiting from the generated data pairs, the proposed method can also enhance Self-Supervised Learning's performance on non-fine-grained visual datasets.

\begin{table}[htb]
  \caption{Top-1 accuracy (in \%) in linear evaluation of our method and MoCo V2 on CIFAR-100 \cite{krizhevsky2009learning} and ImageNet-100 \cite{imagenet_cvpr09}.
  }
  \label{tab:non-fine-grained}
  \centering
  \scalebox{1.0}{
  \setlength{\tabcolsep}{3.0mm}{
  \begin{tabular}{l|c|c}
    \toprule
    
    Methods & CIFAR-100 & ImageNet-100 \\
    \midrule
    MoCo V2 \cite{chen2020improved}        & 61.79      & 62.45      \\
    Ours                                   & \bf{63.26} & \bf{64.02} \\
    
  \bottomrule
\end{tabular}}}
\end{table}

\section{Ablation Studies}
\subsection{The weight $\nu$ of the proposed loss function}
In the overall loss function of our method $\mathcal{L}=\mathcal{L}_C+\alpha\cdot\mathcal{L}_R+\nu\cdot\mathcal{L}_{C_p}$, the weight $\nu$ controls the impact of the proposed $\mathcal{L}_{C_p}$ on the overall performance. To elucidate the effect of varying $\nu$, we conducted experiments on the \texttt{CUB-200} dataset using different values of $\nu$ within our framework. As shown in \cref{tab:ablation-nu}, we achieve the best result when setting $\nu$ to 0.5. Therefore, we set $\nu=0.5$ when implementing our method in all experiments.

\begin{table}[htb]
  \caption{Top-1 accuracy (in \%) in linear evaluation on CUB-200 of our method trained with different $\nu$ values.
  }
  \label{tab:ablation-nu}
  \centering
  \scalebox{1.0}{
  \setlength{\tabcolsep}{3.0mm}{
  \begin{tabular}{l|c|c|c|c}
    \toprule
    
    $\nu$ & 0.0 & 0.1 & 0.5 & 1.0 \\
    \midrule
    Ours  & 64.33 & 65.67 & 66.17 & 65.82\\
    
  \bottomrule
\end{tabular}}}
\end{table}

\subsection{The threshold $\kappa$ in low-variance feature identification criterion}

We explore the impact of the hyperparameter $\kappa$ on the efficacy of feature identification by varying $\kappa$ within our method. This investigation is conducted on the CUB-200 dataset, holding all other experimental settings constant while only adjusting $\kappa$. The results, presented in Table \ref{tab:ablation-kappa}, reveal that optimal performance is attained with $\kappa$ set to 0.02. 

\begin{table}[htb]
  \caption{Top-1 accuracy (in \%) in linear evaluation on CUB-200 of our method trained with different $\kappa$ values in low-variance dimensions identification.
  }
  \label{tab:ablation-kappa}
  \centering
  \scalebox{1.0}{
  \setlength{\tabcolsep}{3.0mm}{
  \begin{tabular}{l|c|c|c|c|c}
    \toprule
    
    $\kappa$ & 0.0 & 0.01 & 0.02 & 0.05 & $+\infty$ \\
    \midrule
    Ours  & 66.04 & 66.10 & 66.17 & 66.15 & 66.05\\
    
  \bottomrule
\end{tabular}}}
\end{table}

\begin{figure}[h]
  \centering
  \includegraphics[width=0.9\textwidth]{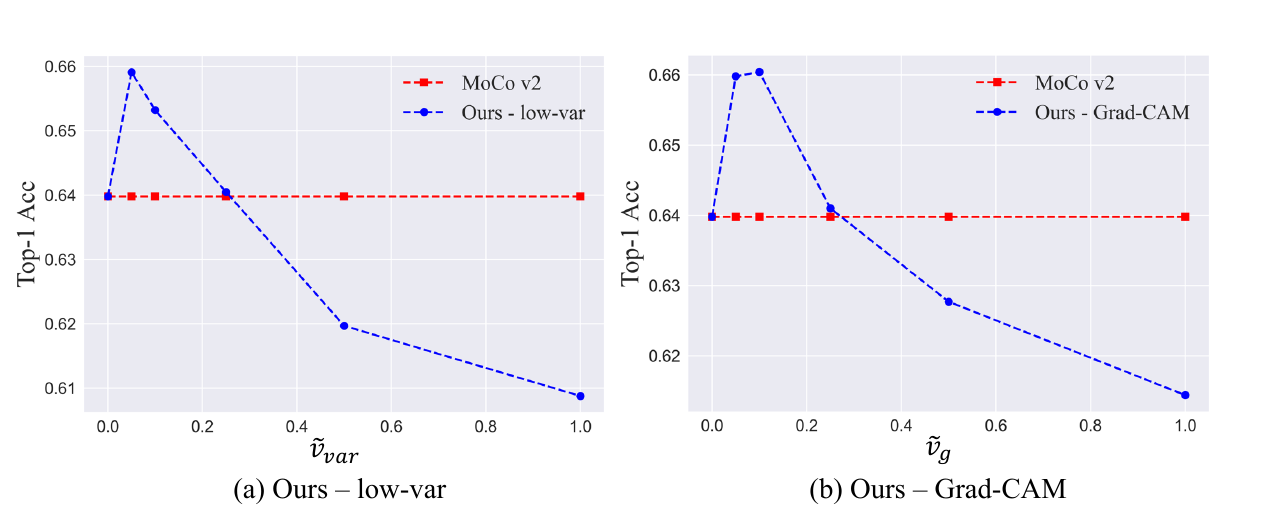}
  \caption{Top-1 accuracy (in \%) in linear evaluation of two configurations of our method, (a) Ours - low-var, (b) Ours - Grad-CAM.
  }
  \label{fig:ablation-noise-std}
\end{figure}

\subsection{The proposed feature identification techniques}

In this work, we proposed to form data pairs between data reconstructed from feature vectors and its perturbed counterparts. Two feature identification techniques are introduced to identify non-discriminative features. To better understand the impact of the feature identification techniques, here we introduce random Gaussian noise to all feature dimensions in a configuration which we term `Ours - random'. As seen in \ref{tab:random-noise}, the two feature identification techniques provide significant performance improvement over this baseline.

\begin{table}[h]
  \caption{Ablation study results on feature identification methods. Top1 linear accuracy is reported. `Ours - Random' denotes a configuration of our method where two feature identification techniques are not used and random Gaussian noise is applied to all feature dimensions.}
  \label{tab:random-noise}
  \centering
  \setlength{\tabcolsep}{8mm}{
  \begin{tabular}{l|ccc}
    \toprule
    Method & CUB-200 & Cars & Aircraft \\
    \midrule
    MoCo V2         & 63.98      & 62.02      & 51.13     \\
    Ours - Random   & 64.12      & 61.84      & 52.09     \\
    Ours            & \bf{66.17} & \bf{65.60} & \bf{55.28}\\
  \bottomrule
\end{tabular}}
\end{table}

\subsection{The intensity of perturbation}

In our approach, we perturb feature vectors by introducing random Gaussian noise with zero mean. The intensity of this noise is governed by the standard deviation of the Gaussian distribution, which in turn is regulated by the hyperparameters $\epsilon_g$ and $\epsilon_{var}$. Consequently, we explore two distinct configurations of our methodology, namely \textbf{Ours - Grad-CAM} and \textbf{Ours - low-var}, varying the hyperparameters $\Tilde{v}_g$ and $\Tilde{v}_{var}$, respectively. As demonstrated in Figure \ref{fig:ablation-noise-std}, for both configurations, performance initially improves with an increase in $\Tilde{v}_g$ and $\Tilde{v}_{var}$, but eventually declines, falling below that of MoCo v2. Minimal perturbation results in images that are too similar to the originals, offering limited new information for the encoder to learn from. Conversely, excessive perturbation risks distorting the original image's identity, and pairs that include such distorted images may introduce an overwhelming amount of noise. Therefore, we finally adopt $\Tilde{v}_g=0.1$ and $\Tilde{v}_{var}=0.05$ for our method.
\end{document}